\documentclass[conference]{IEEEtran}
\IEEEoverridecommandlockouts

\usepackage{amsmath, amssymb, amsthm, amsfonts}
\usepackage{enumitem}
\usepackage{dblfloatfix}

\theoremstyle{plain}

\usepackage{graphicx}
\usepackage{multirow}
\usepackage{balance}
\usepackage{booktabs}
\usepackage[hidelinks]{hyperref}

\usepackage{algorithm}
\usepackage{algorithmic}
\usepackage{threeparttable}

\begin{document}

\title{Explainable Heterogeneous Anomaly Detection in Financial Networks via Adaptive Expert Routing}

\author{
 \IEEEauthorblockN{Zan Li, Rui Fan}
 \IEEEauthorblockA{Rensselaer Polytechnic Institute\\
 Troy, NY, USA\\
 liz37@rpi.edu, fanr09@gmail.com}
 }

\maketitle
\begin{abstract}
Financial anomalies arise from heterogeneous mechanisms---price shocks,
liquidity freezes, contagion cascades, and momentum reversals---yet existing
detectors produce uniform anomaly scores without revealing which mechanism is
failing or where risks concentrate. This hinders targeted responses: liquidity
freezes call for market-making support, whereas price shocks from information
asymmetry call for circuit breakers.
Three key challenges remain unresolved: (1) static graph structures cannot
adapt when correlations shift across regimes; (2) uniform detectors overlook
heterogeneous anomaly signatures; and (3) black-box scores provide no actionable
guidance on which mechanism drives the anomaly.
We address these challenges with an adaptive graph learning framework that
embeds interpretability architecturally rather than post hoc. The framework
constructs stress-modulated graphs that adaptively interpolate between known
sector and geographic relationships and data-driven correlations as market
conditions evolve. Anomalies are decomposed via four mechanism-specific
experts---Price-Shock, Liquidity, Systemic-Contagion, and
Momentum-Reversal---whose routing weights serve as interpretable proxies for
mechanism attribution. A hierarchical Market Pressure Index aggregates
entity-level anomaly scores into graduated market-wide alerts.
On 100 U.S.\ equities (2017--2024), the framework detects all six major market
stress events with a 3.7-day mean lead time, outperforming the strongest
baselines by +33 percentage points in detection rate (AUC 0.888, AP 0.626).
Case studies on the SVB collapse (March 2023) and Japan carry-trade unwind
(August 2024) demonstrate that routing weights automatically distinguish
localized sector-specific crises from systemic multi-sector
propagation---without labeled supervision.
\end{abstract}


\section{Introduction}
\label{sec:intro}

The March 2023 banking crisis exemplifies financial system fragility: 
Silicon Valley Bank (SVB) collapsed within 48 hours, triggering \$42B in 
single-day withdrawal attempts and contagion to Signature Bank~\cite{barr2023svb}. 
In August 2024, Japan's carry-trade unwind sparked a 12\% TOPIX 
crash with severe global spillovers, including sharp sell-offs 
across U.S. equity markets~\cite{aquilina2024carry}. Could early-warning systems have 
detected these crises in advance---and more critically, discerned their 
fundamental nature: the SVB collapse as a \textit{localized} banking-sector 
shock versus the Japan unwind as a \textit{systemic} global contagion? Yet 
existing anomaly detectors remain largely \textit{black 
boxes}~\cite{jin2024survey,cheng2025gnnfraud}, offering little insight into 
where risks emerge, which mechanisms drive them, or how such stresses evolve 
across markets.

\noindent\textbf{The fundamental problem: Detecting anomalies without 
identifying underlying mechanisms provides no actionable guidance.} 
Existing detectors---including temporal models (LSTM-AE~\cite{malhotra2016lstm}, 
TranAD~\cite{tuli2022tranad}), static graphs (DOMINANT~\cite{ding2019deep}, 
CoLA~\cite{liu2021cola}), and dynamic methods (EvolveGCN~\cite{pareja2020evolvegcn}, 
DySAT~\cite{sankar2020dysat}, ROLAND~\cite{you2022roland})---produce only scalar 
anomaly scores indicating \textit{that} failures occur, but not \textit{why}. 
Consider two stocks with identical anomaly scores (0.95): one exhibits a 
bid-ask spread explosion with stable prices---a \textit{liquidity freeze} 
demanding market-making intervention~\cite{amihud2002illiquidity,kirilenko2017flash}; 
the other shows extreme return kurtosis with stable spreads---a \textit{price 
shock} driven by information asymmetry requiring circuit 
breakers~\cite{cont2001empirical,baker2020covid}. Without mechanism attribution, 
identical scores yield indistinguishable alerts despite requiring fundamentally 
different interventions. Distinguishing propagating crises---like SVB's 
contagion to First Republic and Signature Bank through shared depositors and 
interbank exposures~\cite{elliott2014financial}---from isolated failures 
requires understanding \textit{mechanisms}, not just \textit{magnitudes}.

\begin{figure}[t]
\centering
\includegraphics[width=\columnwidth]{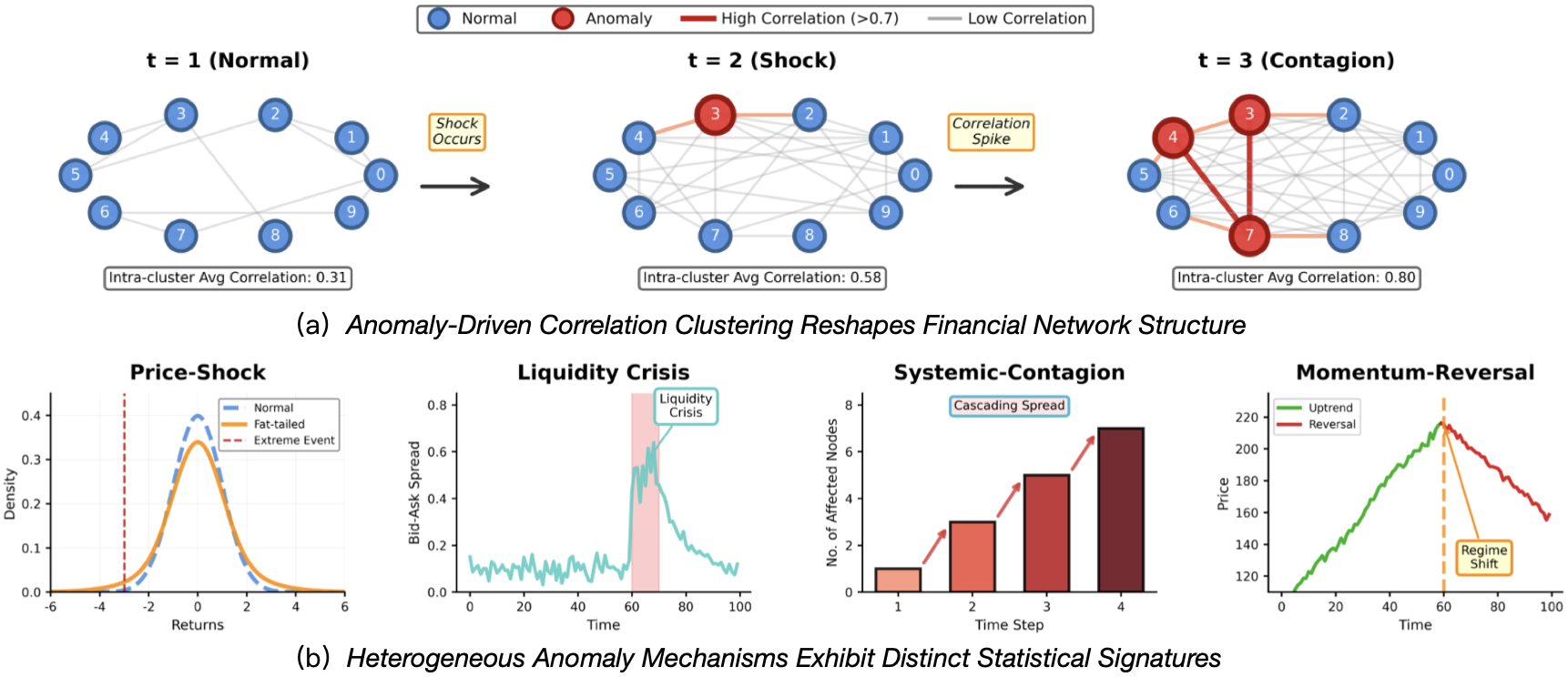}
\caption{\textbf{Three unresolved challenges in financial anomaly detection.}
(a)~Financial networks restructure under stress: intra-cluster correlation
surges from 0.31 to 0.80 during banking distress, yet static graph methods
use a fixed adjacency matrix throughout (\textit{Challenge~1: adaptivity}).
(b)~Four mechanisms produce statistically distinct signatures---fat-tailed
returns (Price-Shock), bid-ask explosion (Liquidity), cascading spread
(Systemic-Contagion), and trend reversal (Momentum-Reversal)---that uniform
detectors collapse into a single scalar score, providing no basis for
identifying which mechanism is active or what intervention is warranted
(\textit{Challenges~2--3: specialization and attribution}).}
\label{fig:problem}
\end{figure}

\noindent\textbf{Our approach: Mechanism-aware detection via specialized 
expert networks with architectural interpretability.} We ground anomaly 
detection in financial economics by identifying four fundamental mechanisms 
that explain \textit{why} market failures occur---Price-Shock, Liquidity, 
Systemic-Contagion, and Momentum-Reversal (Figure~\ref{fig:problem})---each 
with distinct signatures demanding specialized detection. This heterogeneity 
motivates our mixture-of-experts architecture with four specialized detectors 
whose routing weights serve as interpretable proxies for mechanism attribution, 
providing built-in interpretability without post-hoc procedures.

\noindent\textbf{Three technical challenges.} Operationalizing 
mechanism-aware detection requires addressing three concrete obstacles, 
illustrated in Figure~\ref{fig:problem}.

\noindent\textbf{Challenge 1: Static graph structures cannot adapt when 
correlations shift during regime changes.} Financial networks exhibit 
regime-dependent structure: during crises, sectoral and geographic clustering 
strengthen; during calm periods, hidden linkages emerge (supply-chain networks, 
implicit exposures). Figure~\ref{fig:problem}(a) shows SVB's sparse tranquil 
correlations transforming into dense banking clusters as distress propagates. 
Static methods (DOMINANT~\cite{ding2019deep}, CoLA~\cite{liu2021cola}) miss 
such structural shifts by relying on fixed adjacency matrices. Dynamic 
approaches face a dilemma: data-driven methods 
(EvolveGCN~\cite{pareja2020evolvegcn}, DySAT~\cite{sankar2020dysat}) often 
destabilize under distribution shifts; topology-constrained methods 
(ROLAND~\cite{you2022roland}) cannot capture emergent contagion pathways when 
independent institutions couple through correlated stress.

\noindent\textbf{Challenge 2: Uniform detectors cannot distinguish 
heterogeneous anomaly mechanisms with distinct signatures.} 
Figure~\ref{fig:problem}(b) illustrates four mechanisms with unique patterns:
\begin{itemize}[leftmargin=*,itemsep=0.5pt]
\item \textbf{Price-Shock}---Information-driven volatility from 
asymmetric news arrival~\cite{cont2001empirical,baker2020covid}, 
manifesting as fat-tailed returns (kurtosis $>$10) with stable 
spreads.
\item \textbf{Liquidity}---Trading frictions from market-maker 
withdrawal~\cite{amihud2002illiquidity,kirilenko2017flash}, 
manifesting as bid-ask spread explosion with minimal price movement.
\item \textbf{Systemic-Contagion}---Cross-market propagation through 
correlation networks~\cite{elliott2014financial,diebold2014network}, 
manifesting as coordinated cross-sector distress with pronounced 
spread widening.
\item \textbf{Momentum-Reversal}---Regime transitions from trend 
exhaustion~\cite{arnott2023momentum} or structural 
breaks~\cite{hamilton1989new}, manifesting as gradual factor loading 
shifts with RSI sign reversals.
\end{itemize}
Existing methods apply uniform feature processing and suffer from 
\textit{mechanism conflation}: temporal 
models~\cite{malhotra2016lstm,tuli2022tranad} treat all features identically, 
missing contagion effects; static graph 
methods~\cite{ding2019deep,deng2021graph} cannot distinguish rapid shocks 
from gradual transitions; hybrid approaches 
(MTAD-GAT~\cite{zhao2020multivariate}) average mechanism-specific signals 
with irrelevant features under uniform reconstruction. High reconstruction 
errors may thus stem from any mechanism, yet uniform scoring provides no 
basis for distinguishing among them. Graph-based MoE 
methods~\cite{huang2025graphmoe} demonstrate the value of expert routing for 
heterogeneous data, but partitioning by sensor-level heterogeneity rather 
than financially-grounded failure modes does not yield routing weights 
interpretable as mechanism attributions---leaving the question of 
\textit{which} financial mechanism is failing, and \textit{what} intervention 
is warranted, unaddressed.

\noindent\textbf{Challenge 3: Black-box anomaly scores provide no actionable 
guidance on underlying mechanisms or their temporal evolution.} The 
interpretability gap is structural: existing detectors---autoencoders, GANs, 
and normalizing flows---produce scalar scores indicating \textit{that} 
anomalies exist, not \textit{which mechanism} fails. While explainability has 
attracted growing attention in financial AI~\cite{yeo2025finxai}, existing 
approaches remain predominantly post-hoc. Post-hoc methods (e.g., 
SHAP~\cite{lundberg2017shap}, attention visualization) face well-documented 
limitations~\cite{turbe2023posthoc}: attributions vary across consecutive 
days, and feature-level statements cannot distinguish liquidity stress from 
contagion. The consequence is regulatory inefficiency: misidentifying solvency 
stress as liquidity problems results in misallocated interventions.

Addressing these challenges in isolation is insufficient: adaptivity 
without specialization still yields black-box scores, and specialization 
without interpretability leaves practitioners guessing which mechanism 
failed. This paper presents a detector that solves all three 
simultaneously, grounded in financial economics.

\noindent\textbf{Contributions.}

\noindent\textbf{1. Unified mechanism-aware framework.} We present the 
first framework to jointly address all three challenges by integrating 
(i)~stress-modulated adaptive graph fusion between domain structural priors 
and learned correlations, (ii)~four financially-grounded mechanism-specific 
experts decomposing anomalies into Price-Shock, Liquidity, 
Systemic-Contagion, and Momentum-Reversal, and (iii)~architectural 
interpretability via routing weight dynamics embedded in the model rather 
than applied post-hoc. No existing method achieves all three simultaneously 
(Table~\ref{tab:comparison}).

\noindent\textbf{2. Demonstrated detection superiority on real financial 
crises.} Evaluated on 100 U.S. equities (2017--2024), the framework achieves 
100\% detection of six major stress events with a 3.7-day mean lead time, 
substantially outperforming representative baselines. A hierarchical Market 
Pressure Index---aggregating entity-level anomaly scores via anomaly rate, 
peak intensity, tail concentration, and score dispersion---provides graduated 
market-wide pressure signals for operationally actionable risk monitoring.

\noindent\textbf{3. Interpretable mechanism attribution without labeled 
supervision.} Unlike post-hoc attribution methods, routing weights 
$\mathbf{w}_{i,t}$ emerge directly from forward inference, providing 
built-in mechanism attribution at no additional cost. Their 
baseline-relative trajectories automatically distinguish localized from 
systemic crises---demonstrated on SVB (March 2023) and Japan carry-trade 
unwind (August 2024)~\cite{aquilina2024carry}---without any crisis 
annotations or labeled supervision.

\noindent\textbf{Organization.} Section~\ref{sec:related} reviews related 
work and positions our contributions. Sections~\ref{sec:formulation} 
and~\ref{sec:method} formalize the problem and present the framework 
architecture. Sections~\ref{sec:experiments} and~\ref{sec:conclusion} 
report empirical validation and concluding remarks.

\section{Related Work}
\label{sec:related}

Our framework sits at the intersection of temporal and graph-based 
anomaly detection, mechanism-aware modeling, and interpretable AI 
in finance. We review each in turn and identify the gaps our work 
addresses.

\subsection{Temporal and Graph-Based Anomaly Detection}

\noindent\textbf{Temporal models.}
Deep temporal approaches capture nonlinear dependencies via unsupervised 
reconstruction: LSTM-AE~\cite{malhotra2016lstm} pioneered encoder--decoder 
architectures for sequence anomaly scoring; TranAD~\cite{tuli2022tranad} 
introduced transformer-based dual attention with adversarial training; 
OmniAnomaly~\cite{su2019robust} combined variational autoencoders with 
stochastic recurrent networks. These methods treat entities 
independently, overlooking cross-entity interactions critical where 
contagion propagates through shared exposures.

\noindent\textbf{Graph-based models.}
DOMINANT~\cite{ding2019deep} and GDN~\cite{deng2021graph} extend graph 
convolutional networks~\cite{kipf2017gcn} to autoencoder architectures, 
but assume static adjacency matrices that cannot adapt when correlations 
shift across regimes. Dynamic extensions address temporal evolution: 
EvolveGCN~\cite{pareja2020evolvegcn} evolves GCN parameters via RNNs; 
DySAT~\cite{sankar2020dysat} applies self-attention over graph snapshots; 
ROLAND~\cite{you2022roland} imposes topological constraints for stability. 
Hybrid approaches (MTAD-GAT~\cite{zhao2020multivariate}, 
CoLA~\cite{liu2021cola}) integrate temporal and graph modalities but employ 
uniform reconstruction objectives---data-driven methods overfit 
panic-driven synchronization during crises while topology-constrained 
methods miss emergent cross-sector contagion 
pathways~\cite{tzagkarakis2024dynamic}.

\subsection{Mechanism-Aware Anomaly Detection}

\noindent\textbf{Financial crisis theory and mechanisms.}
Financial economics identifies four canonical failure modes---price 
shocks, liquidity crises, systemic contagion, and momentum 
reversals~\cite{cont2001empirical,amihud2002illiquidity,elliott2014financial,arnott2023momentum}---yet 
modern deep models output scalar scores without mechanism attribution, 
offering no guidance on which intervention is warranted.

\noindent\textbf{Mixture-of-experts for specialized detection.}
Mixture-of-Experts (MoE)~\cite{jacobs1991adaptive} decomposes tasks into 
specialized subproblems via gating. In anomaly detection, graph-based MoE 
with memory-augmented routers demonstrates the value of expert routing for 
heterogeneous industrial time series~\cite{huang2025graphmoe}. However, 
existing uses partition the input space by sensor-level heterogeneity 
rather than financially-grounded failure modes---routing weights thus carry 
no economically meaningful mechanism attribution. Our experts are instead 
grounded in financial economics, making routing weights economically 
interpretable mechanism attributions.

\subsection{Interpretability: Architectural vs.\ Post-Hoc}

Interpretability has attracted growing attention in financial 
AI~\cite{yeo2025finxai}, with post-hoc methods---GNNExplainer~\cite{ying2019gnnexplainer} 
and SHAP~\cite{lundberg2017shap}---dominating current practice. However, 
systematic evaluation of post-hoc attribution 
methods~\cite{turbe2023posthoc} demonstrates temporal instability and 
feature-level granularity that limit diagnostic value: attributions vary 
across consecutive days, and feature-level statements cannot distinguish 
mechanism types. Financial anomaly detection thus lacks methods 
that embed mechanism-level attribution \textit{within} the detection 
architecture itself.

\subsection{Positioning and Novelty}

Table~\ref{tab:comparison} contrasts representative methods along five 
dimensions. Our framework is the first to jointly achieve stress-modulated 
adaptive graphs, mechanism-aligned expert specialization, and architectural 
interpretability---no existing method achieves all three simultaneously. 
Together these capabilities translate detection into actionable guidance 
aligned with documented crisis transmission channels.

\begin{table}[t]
\centering
\caption{\textbf{Comparison with representative approaches.} Our framework 
uniquely combines stress-modulated adaptive graphs, mechanism-aligned 
experts, and architectural interpretability.}
\label{tab:comparison}
\resizebox{\columnwidth}{!}{
\begin{tabular}{lccccc}
\toprule
\textbf{Method} & \textbf{Temporal} & \textbf{Graph} & \textbf{Adaptive} 
& \textbf{Specialized} & \textbf{Interpretable} \\
\midrule
LSTM-AE~\cite{malhotra2016lstm}      & \checkmark & $\times$   
& $\times$            & $\times$   & $\times$ \\
TranAD~\cite{tuli2022tranad}         & \checkmark & $\times$   
& $\times$            & $\times$   & $\times$ \\
DOMINANT~\cite{ding2019deep}         & $\times$   & \checkmark 
& $\times$            & $\times$   & $\times$ \\
EvolveGCN~\cite{pareja2020evolvegcn} & \checkmark & \checkmark 
& Unstable            & $\times$   & $\times$ \\
ROLAND~\cite{you2022roland}          & \checkmark & \checkmark 
& Limited             & $\times$   & $\times$ \\
MTAD-GAT~\cite{zhao2020multivariate} & \checkmark & \checkmark 
& $\times$            & $\times$   & $\times$ \\
\midrule
\textbf{Ours} & \checkmark & \checkmark & \textbf{Stress-Mod} 
& \checkmark & \textbf{Arch.} \\
\bottomrule
\end{tabular}
}
\\[4pt]
{\footnotesize\raggedright
\textit{Adaptive}: $\times$~=~static graph; Unstable~=~destabilizes under 
distribution shift; Limited~=~fixed topology; Stress-Mod~=~stress-modulated 
adaptive rebalancing. \textit{Arch.}~=~architectural (during-inference) 
interpretability.\par}
\end{table}

\section{Problem Formulation}
\label{sec:formulation}

We formalize financial anomaly detection as \textbf{mechanism-aware 
inference} on temporal dynamic graphs, unifying three goals: 
(1)~\textit{localization}---identifying which entities exhibit anomalies 
via node-level scores; (2)~\textit{attribution}---explaining why anomalies 
occur through mechanism-specific routing weights; and 
(3)~\textit{systemic quantification}---assessing market-wide risks through 
a network-level pressure index.

\subsection{Input: Temporal Dynamic Financial Networks}

\noindent\textbf{Network structure.} A financial system with $N$ stocks is 
represented as a temporal graph $\mathcal{G} = (\mathcal{V}, 
\{\mathcal{E}_t\}, \mathbf{A}_{\text{prior}})$, where $\mathcal{V}$ is the 
set of stocks ($|\mathcal{V}| = N$), $\mathcal{E}_t$ denotes time-varying 
correlations, and $\mathbf{A}_{\text{prior}} \in [0,1]^{N \times N}$ 
encodes stable sectoral and geographic domain knowledge: 
$\mathbf{A}_{\text{prior}}[i,j] > 0$ if stocks $i$ and $j$ share industry 
(GICS) or geographic region. This prior serves as the baseline topology for 
stress-modulated fusion (Section~\ref{sec:module2}).

\noindent\textbf{Node features and temporal windows.} Each stock $i \in 
\mathcal{V}$ has multivariate features $\mathbf{X}_{i,t} \in \mathbb{R}^{T 
\times F}$, where $T = 20$ (approximately one month of daily data) and $F = 
29$. At each time $t$, we process the window $[t - T + 1, t]$.

\noindent\textbf{Mechanism-specific feature subsets.} The $F = 29$ features 
are partitioned into four mechanism-aligned subsets:
\begin{itemize}[leftmargin=*,itemsep=0.5pt]
\item \textbf{Price-Shock} (6 features): volatility and return 
  distribution~\cite{cont2001empirical}
\item \textbf{Liquidity} (8 features): market microstructure and 
  market-maker activity~\cite{amihud2002illiquidity}
\item \textbf{Systemic-Contagion} (7 features): cross-sectional correlation 
  and spillover~\cite{elliott2014financial,diebold2014network}
\item \textbf{Momentum-Reversal} (8 features): technical indicators for 
  regime transitions~\cite{arnott2023momentum,hamilton1989new}
\end{itemize}

\subsection{Output: Interpretable Multi-Level Detection}
\label{sec:output}

At each time $t$, the framework produces three outputs:

\noindent\textbf{(1) Entity-level anomaly scores:} $s_{i,t} \in [0,1]$ 
quantify deviation intensity for each stock, paired with mechanism 
attribution via routing weights.

\noindent\textbf{(2) Routing weights for mechanism attribution:}
$\mathbf{w}_{i,t} = [w_{i,t}^{(1)}, w_{i,t}^{(2)}, w_{i,t}^{(3)}, 
w_{i,t}^{(4)}]^\top \in \mathbb{R}^4$ with $\sum_k w_{i,t}^{(k)} = 1$ 
and $w_{i,t}^{(k)} \geq 0$, produced during forward inference as 
interpretable proxies for mechanism attribution. Shannon entropy quantifies 
mechanism concentration:
\begin{equation}
H(\mathbf{w}_{i,t}) = -\sum_{k=1}^{4} w_{i,t}^{(k)} \ln w_{i,t}^{(k)}
\label{eq:routing_entropy}
\end{equation}
where $H_{\min} = 0$ indicates single-mechanism dominance and $H_{\max} = 
\ln 4 \approx 1.386$ indicates uniform distribution. Low entropy signals 
localized, targetable anomalies; high entropy signals coordinated 
multi-mechanism stress.

\noindent\textbf{(3) Market Pressure Index (MPI):} $\text{MPI}_t \in 
[0,1]$ aggregates entity-level anomalies into a systemic risk indicator 
with four hierarchical alert levels (L1--L4) for graduated escalation 
aligned with regulatory protocols. The full MPI formula is detailed in 
Section~\ref{sec:method}.

\subsection{Statistical Proxy Labels for Validation}

Since the framework is fully unsupervised, proxy labels derived from 
mechanism-specific indicators at the 95th percentile support hyperparameter 
validation:
\begin{itemize}[leftmargin=*,itemsep=0.5pt]
\item \textbf{Price-Shock}: extreme returns~\cite{cont2001empirical}
\item \textbf{Liquidity}: Amihud 
  illiquidity~\cite{amihud2002illiquidity}
\item \textbf{Systemic-Contagion}: market 
  correlation~\cite{elliott2014financial,diebold2014network}
\item \textbf{Momentum-Reversal}: RSI 
  extremes~\cite{arnott2023momentum,hamilton1989new}
\end{itemize}
These proxy labels are used only for hyperparameter selection (AUC, AP on 
validation data) and never affect model training or threshold selection. 
Proxy labels measure cross-sectional extremity at each time $t$, whereas 
anomaly scores measure entity-specific reconstruction error relative to 
each stock's learned normal baseline---the two constructs are 
informationally distinct despite sharing underlying financial variables.

\section{Method}
\label{sec:method}

\begin{figure}[t]
\centering
\includegraphics[width=\columnwidth]{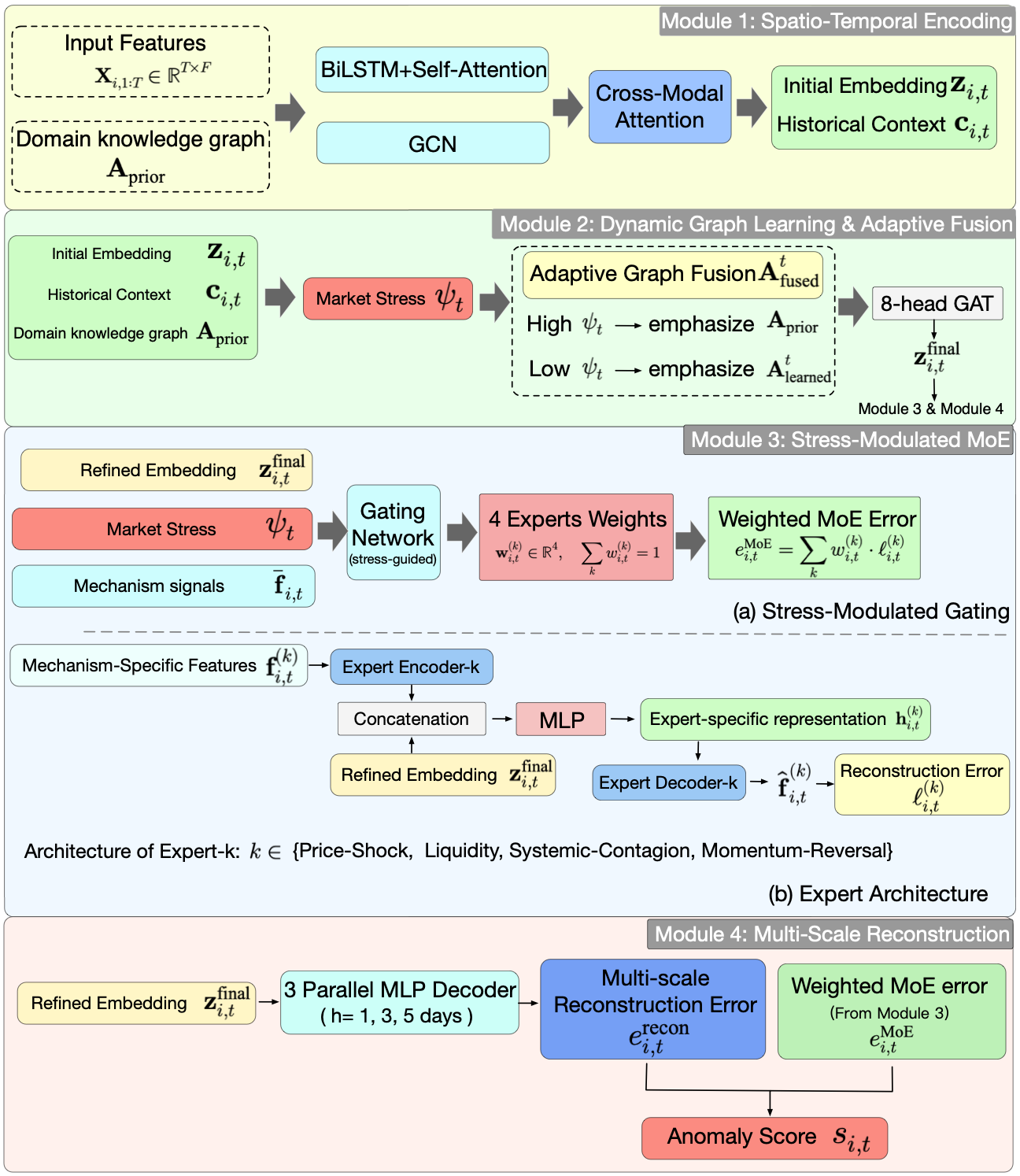}
\caption{\textbf{Mechanism-aware framework for financial anomaly detection.}
Module~1 encodes temporal dynamics and cross-stock dependencies.
Module~2 constructs a stress-adaptive graph that shifts between structural
priors and learned correlations as market conditions change.
Module~3 routes each entity to four mechanism-specific experts (panel~a);
panel~b shows individual expert architecture.
Module~4 aggregates expert outputs into entity anomaly scores and a
hierarchical Market Pressure Index.}
\label{fig:architecture}
\end{figure}

\noindent\textbf{Notation.} We write $\operatorname{sigmoid}(\cdot)$ 
explicitly to avoid ambiguity with standard deviation $\sigma$. The 
framework processes windows $\mathbf{X}_{1:N,t-T+1:t} \in 
\mathbb{R}^{N \times T \times F}$ and outputs entity anomaly scores 
$\{s_{i,t}\}$, mechanism weights $\{\mathbf{w}_{i,t}\}$, and 
$\text{MPI}_t$ via four modules: spatial-temporal encoding (Module~1), 
stress-modulated graph fusion (Module~2), mechanism-specific routing 
(Module~3), and multi-scale aggregation (Module~4). 
Figure~\ref{fig:architecture} illustrates the architecture.

\subsection{Module 1: Spatial-Temporal Encoding}
\label{sec:module1}

\noindent\textbf{Temporal branch.} BiLSTM processes input 
$\mathbf{X}_{i,1:T}$, concatenating forward and backward hidden states 
(each 128-dim) to form $\mathbf{h}_{i,t}^{\text{bi}} \in \mathbb{R}^{T 
\times 256}$. Multi-head self-attention (4 heads) then captures long-range 
temporal dependencies, yielding $\mathbf{h}_{i,t}^{\text{attn}} \in 
\mathbb{R}^{T \times 256}$. Mean pooling and final-step output are 
concatenated and projected to $\mathbf{h}_{i,t}^{\text{temp}} \in 
\mathbb{R}^{128}$.

\noindent\textbf{Spatial branch.} The feature window is projected to 
$\mathbf{h}_i^{(0)} \in \mathbb{R}^{128}$. Two GCN 
layers~\cite{kipf2017gcn} propagate information over 
$\mathbf{A}_{\text{prior}}$:
\begin{equation}
\mathbf{h}_i^{(\ell)} = \text{ReLU}\!\left(\tilde{\mathbf{A}}_{\text{prior}}
\mathbf{h}_i^{(\ell-1)} \mathbf{W}_{\text{gcn}}^{(\ell)}\right), 
\quad \ell \in \{1,2\}
\end{equation}
where $\tilde{\mathbf{A}}_{\text{prior}} = \mathbf{D}^{-1}
(\mathbf{A}_{\text{prior}} + \mathbf{I})$ adds self-loops and 
row-normalizes. A residual connection 
$\mathbf{h}_i^{(\ell)} \leftarrow \mathbf{h}_i^{(\ell-1)} + 
0.5\,\mathbf{h}_i^{(\ell)}$ yields 
$\mathbf{h}_{i,t}^{\text{spat}} \in \mathbb{R}^{128}$.

\noindent\textbf{Cross-modal fusion and output.} Cross-attention fuses 
temporal (query) and spatial (key/value) representations with a residual 
skip, yielding $\mathbf{h}_{i,t}^{\text{fused}} \in \mathbb{R}^{128}$. 
Two-layer MLPs with LayerNorm and ReLU then produce an initial embedding 
$\mathbf{z}_{i,t} \in \mathbb{R}^{128}$ and historical context 
$\mathbf{c}_{i,t} \in \mathbb{R}^{64}$.

\subsection{Module 2: Stress-Modulated Adaptive Graph Learning}
\label{sec:module2}

\noindent\textbf{Market stress.} Four financial indicators are 
aggregated into a unified stress measure $\psi_t \in [0,1]$:
\begin{equation}
\psi_t = \sum_{k=1}^4 \beta_k \tilde{I}_k^t, \qquad
\tilde{I}_k^t = \frac{I_k^t - \min_\tau I_k^\tau}
{\max_\tau I_k^\tau - \min_\tau I_k^\tau}
\label{eq:market_stress}
\end{equation}
where $\beta_k$ are learnable softmax-normalized weights. Raw 
indicators:
\begin{align}
I_1^t &= \mathrm{std}_{i,\tau}(r_{i,\tau})_{\tau \in [t-T+1,t]}
\quad \text{(return volatility)} \nonumber \\[2pt]
I_2^t &= \tfrac{2}{N(N-1)} \sum_{i<j} 
\mathrm{corr}(r_{i,t-T+1:t}, r_{j,t-T+1:t})
\quad \text{(avg correlation)} \nonumber \\[2pt]
I_3^t &= \tfrac{1}{NT} \sum_{i=1}^N \sum_{\tau=t-T+1}^{t}
\mathbb{1}\bigl[|r_{i,\tau}| > 2\sigma_{r,\tau}\bigr]
\nonumber \\
&\qquad\qquad\qquad\qquad\qquad
\text{(extreme return fraction)} \nonumber \\[2pt]
I_4^t &= \tfrac{1}{8NT} \sum_{i} \sum_{j=1}^{8} 
\sum_{\tau=1}^{T} \mathbf{f}_{i,\tau}^{(2)}[j]
\quad \text{(liquidity intensity)} \nonumber
\end{align}

\noindent\textbf{Multi-source graph construction.} The learned graph 
combines three sources with learnable softmax-normalized weights:
\begin{equation}
\mathbf{A}_{\text{learned}}^t = w_{\text{temp}}\mathbf{A}_{\text{temp}}^t 
+ w_{\text{ctx}}\mathbf{A}_{\text{ctx}}^t + 
w_{\text{prior}}\mathbf{A}_{\text{prior}}
\end{equation}
where temporal and contextual similarities are cosine similarities of 
$\mathbf{z}_{i,t}$ and $\mathbf{c}_{i,t}$ respectively. Top-$k$ 
sparsification ($k=20$) is applied via a differentiable soft-threshold:
\begin{equation}
M[i,j] = \operatorname{sigmoid}\!\left(5(\mathbf{A}_{\text{learned}}^t
[i,j] - \theta_{(k)}^i)\right)
\end{equation}
where $\theta_{(k)}^i$ is the $k$-th largest value in row $i$. Symmetric 
masking and row-wise softmax normalization are then applied.

\noindent\textbf{Stress-modulated fusion.} The fusion coefficient $\alpha_t$ 
is modulated by market stress:
\begin{equation}
\alpha_t = \text{clamp}\!\left(\operatorname{sigmoid}(\alpha_{\text{base}} 
+ \beta_\alpha\psi_t),\; 0.2,\; 0.8\right)
\label{eq:stress_weight}
\end{equation}
\begin{equation}
\mathbf{A}_{\text{fused}}^t = \alpha_t\mathbf{A}_{\text{prior}} +
(1-\alpha_t)\mathbf{A}_{\text{learned}}^t
\label{eq:fused_graph}
\end{equation}
High stress ($\psi_t \to 1$) drives $\alpha_t \to 0.8$ (emphasizing 
sectoral/geographic clustering); low stress drives $\alpha_t \to 0.2$ 
(favoring learned correlations). Clamping to $[0.2, 0.8]$ ensures both 
sources always contribute.

\noindent\textbf{Graph attention refinement.} An 8-head 
GAT~\cite{velickovic2018gat} refines embeddings over 
$\mathbf{A}_{\text{fused}}^t$:
\begin{equation}
\mathbf{z}_{i,t}^{\text{final}} = \mathbf{z}_{i,t} + 0.5\sum_{j \in 
\mathcal{N}_i^t} a_{ij}\mathbf{W}_{\text{gat}}\mathbf{z}_{j,t} \in 
\mathbb{R}^{128}
\end{equation}

\subsection{Module 3: Stress-Modulated Mixture-of-Experts}
\label{sec:module3}

The MoE paradigm assigns each entity to a combination of four specialist 
networks, each responsible for one anomaly mechanism. Routing weights 
$\mathbf{w}_{i,t}$ are learned end-to-end and serve as interpretable 
proxies for mechanism attribution.

\noindent\textbf{Feature partitioning.} The $F=29$ features are partitioned 
into four mechanism-specific subsets as defined in 
Section~\ref{sec:formulation}.

\noindent\textbf{Mechanism signal extraction.} For each mechanism $k$, 
the mean absolute feature value $\bar{f}_{i,t}^{(k)} = 
\frac{1}{d_k}\sum_{j=1}^{d_k}|\mathbf{f}_{i,T}^{(k)}[j]|$, where 
$d_k \in \{6,8,7,8\}$, serves as an inductive prior for routing, 
grounding gating in observable mechanism-specific activity.

\noindent\textbf{Stress-modulated gating.} Stock $i$ attends over all $N$ 
peers via cross-entity attention~\cite{vaswani2017attention}:
\begin{equation}
\mathbf{a}_{i,t} = \text{MultiHeadAttn}(\mathbf{z}_{i,t}^{\text{final}},\;
\mathbf{Z}^{\text{final}},\; \mathbf{Z}^{\text{final}})
\end{equation}
Gating logits are computed as:
\begin{equation}
\mathbf{g}_{i,t} = \text{MLP}\!\left([\mathbf{z}_{i,t}^{\text{final}};
\mathbf{a}_{i,t}; \psi_t; \bar{\mathbf{f}}_{i,t}]\right) + 
\mathbf{b}_{\text{div}} \in \mathbb{R}^4
\end{equation}
where $\mathbf{b}_{\text{div}} = [0, 0.5, 0.3, 0.2]$ provides an initial 
prior favoring Liquidity and Contagion experts. Temperature is 
stress-modulated:
\begin{equation}
\tau_t = \text{clamp}(|\tau_{\text{base}}| + \beta_{\text{stress}}\psi_t,\; 
0.5,\; 3.0)
\end{equation}
\begin{equation}
\mathbf{w}_{i,t} = \text{Softmax}(\mathbf{g}_{i,t}/\tau_t)
\label{eq:routing_weights}
\end{equation}
High stress increases $\tau_t$ for softer multi-mechanism routing; low 
stress sharpens routing toward the dominant mechanism. During training, 
Gaussian noise and a weight floor $\delta=0.1$ ensure all experts 
receive sufficient training signal.

\noindent\textbf{Expert architecture.} Each expert $k$ encodes 
its feature subset via $\text{FeatureEncoder}_k$ 
($\mathbb{R}^{d_k} \to \mathbb{R}^{64}$), fuses with global 
context $\mathbf{u}_{i,t}^{(k)}$ via $\text{FusionMLP}_k$ 
($\to \mathbb{R}^{128}$), and reconstructs via 
$\text{Decoder}_k$ ($\to \mathbb{R}^{d_k}$), where 
$\mathbf{u}_{i,t}^{(k)} = \mathbf{z}_{i,t}^{\text{GAT}}$ for 
the Systemic-Contagion expert and 
$\mathbf{z}_{i,t}^{\text{final}}$ for the others. The mixture 
error is:
\begin{equation}
e_{i,t}^{\text{MoE}} = \sum_{k=1}^4 w_{i,t}^{(k)} \cdot 
\frac{1}{d_k}\|\mathbf{f}_{i,T}^{(k)} - 
\hat{\mathbf{f}}_{i,t}^{(k)}\|_2^2
\end{equation}

\subsection{Module 4: Multi-Scale Aggregation}
\label{sec:module4}

\noindent\textbf{Multi-scale reconstruction.} Three parallel MLP decoders 
reconstruct feature history at horizons $h \in \{1,3,5\}$ days, combined 
with short-term signals weighted most heavily:
\begin{equation}
e_{i,t}^{\text{recon}} = 0.5\,e_{i,t}^{(1)} + 0.3\,e_{i,t}^{(3)} +
0.2\,e_{i,t}^{(5)}
\end{equation}

\noindent\textbf{Entity-level anomaly score.} MoE and reconstruction errors 
are combined and batch-standardized:
\begin{equation}
s_{i,t} = \operatorname{sigmoid}\!\left(2 \cdot 
\frac{0.6\,e_{i,t}^{\text{MoE}} + 0.4\,e_{i,t}^{\text{recon}} - 
\mu_e^{(b)}}{\sigma_e^{(b)} + \epsilon}\right)
\label{eq:entity_score}
\end{equation}

\noindent\textbf{Market Pressure Index.} $\text{MPI}_t$ aggregates four 
normalized indicators:
\begin{equation}
\text{MPI}_t = 0.30\,m_1^t + 0.20\,m_2^t + 0.30\,m_3^t + 0.20\,m_4^t
\end{equation}
\begin{align}
m_1^t &= \operatorname{sigmoid}\!\left(\tfrac{\bar{s}_t - 
\mu_{\text{mean}}}{\sigma_{\text{mean}} + \epsilon}\right)
\quad \text{(mean anomaly rate)} \\
m_2^t &= \operatorname{sigmoid}\!\left(\tfrac{\text{std}_i(s_{i,t}) - 
\mu_{\text{disp}}}{\sigma_{\text{disp}} + \epsilon}\right)
\quad \text{(cross-sectional dispersion)} \\
m_3^t &= \operatorname{sigmoid}\!\left(\tfrac{\frac{1}{N}\sum_i
\mathbb{1}[s_{i,t}>P_{90}] - \mu_{\text{tail}}}{\sigma_{\text{tail}} + 
\epsilon}\right) \quad \text{(tail concentration)} \\
m_4^t &= \operatorname{sigmoid}\!\left(\tfrac{\max_i(s_{i,t}) - 
\mu_{\text{peak}}}{\sigma_{\text{peak}} + \epsilon}\right)
\quad \text{(peak intensity)}
\end{align}
The higher weight on $m_3$ reflects the empirical pattern that extreme 
anomalies in few entities precede widespread 
contagion~\cite{elliott2014financial}. All baseline statistics and $P_{90}$ 
are computed once from the first 100 training days and permanently fixed, 
enabling deployment without recalibration.

\noindent\textbf{Hierarchical alerts.} Four alert levels from 
validation-set percentiles:
\begin{itemize}[leftmargin=*,itemsep=0.5pt]
\item L1 (Observation): $\text{MPI}_t \geq P_{70}$
\item L2 (Attention): $\text{MPI}_t \geq P_{85}$
\item L3 (Warning): $\text{MPI}_t \geq P_{95}$
\item L4 (Crisis): $\text{MPI}_t \geq P_{99}$
\end{itemize}

\subsection{Training and Inference}

\noindent\textbf{Loss function.} The framework trains end-to-end with an 
unsupervised objective:
\begin{equation}
\mathcal{L} = 0.4\,\mathcal{L}_{\text{MoE}} + 0.3\,\mathcal{L}_{\text{rec}}
+ 0.1\,\mathcal{L}_{\text{div}} + \lambda_{\text{reg}}\,
\mathcal{L}_{\text{reg}} + 0.05\,\mathcal{L}_{\text{aux}}
\end{equation}
$\mathcal{L}_{\text{MoE}}$ and $\mathcal{L}_{\text{rec}}$ minimize 
reconstruction error across mechanism experts and temporal horizons. 
$\mathcal{L}_{\text{div}}$ promotes expert specialization via three 
terms: (i)~an entropy floor below $0.8\,H_{\max}$; (ii)~a collapse 
penalty bounding batch-mean routing weight $\bar{w}_k$ to $[0.15, 
0.35]$; and (iii)~an error diversity term penalizing cosine similarity 
between expert reconstruction error profiles. $\mathcal{L}_{\text{reg}}$ 
is lightweight L2 regularization ($\lambda_{\text{reg}} \approx 
5\times10^{-5}$). $\mathcal{L}_{\text{aux}}$ (weight 0.05) comprises a 
graph sparsity penalty, a spatial-temporal balance loss, and a graph 
prior consistency penalty.

\noindent\textbf{Optimization.} AdamW with $\eta = 5\times10^{-4}$ (main) 
and $\eta_{\text{graph}} = 2.5\times10^{-3}$ (graph learner), cosine 
annealing ($T_{\max}=50$), batch size 32, gradient clipping at norm 1.0, 
early stopping with patience 10.

\noindent\textbf{Inference and mechanism attribution.} Routing weights 
$\mathbf{w}_{i,t}$ from forward inference provide immediate mechanism 
identification: low $H(\mathbf{w}_{i,t})$ indicates single-mechanism 
stress; high entropy indicates coordinated multi-mechanism anomalies. 
For case studies, weights are post-processed into baseline-relative 
changes against pre-event windows (30--20 days before event 
onset)~\cite{mackinlay1997event}, for visualization only. Inference 
runs at 50ms per timestep on NVIDIA A100 ($N=100$); model size: 1.85M 
parameters.

\section{Experiments}
\label{sec:experiments}

\subsection{Experimental Setup}

\noindent\textbf{Dataset.} We use 100 U.S. equities drawn from S\&P 500 
constituents as of January 2017, with balanced sector representation 
across all 11 GICS sectors. Fixing the constituent list at January 2017 
avoids look-ahead bias; requiring continuous listing through 2017--2024 
introduces a modest survivorship bias, with limited impact on 
event-level detection since all six test events are systemic or 
sector-wide rather than idiosyncratic firm failures. Data are sourced 
from WRDS. Temporal splits: Train 2017--2021 (1,240 days), Validation 
2022 (232 days), Test 2023--2024 (483 days).

\noindent\textbf{Test events.} Six major stress events identified prior to 
model training from established financial literature and regulatory 
reports~\cite{barr2023svb,aquilina2024carry}: SVB Collapse (March 10, 
2023), Signature Bank Failure (March 13, 2023), Credit Suisse Crisis 
(March 20, 2023), Tech Earnings Selloff (October 27, 2023), Weak Jobs 
Report (August 2, 2024), and Japan Carry-Trade Unwind (August 5, 2024). 
Planned monetary policy announcements are excluded as predictable. The 
three March 2023 banking events are treated as distinct episodes, each 
with a different institution and resolution mechanism (depositor run, 
FDIC closure, AT1 bond writedown).

\noindent\textbf{Baselines.} Temporal: LSTM-AE~\cite{malhotra2016lstm}, 
TranAD~\cite{tuli2022tranad}, OmniAnomaly~\cite{su2019robust}. Static 
graph: DOMINANT~\cite{ding2019deep}, GDN~\cite{deng2021graph}, 
AnomalyDAE~\cite{fan2020anomalydae}. Dynamic graph: 
EvolveGCN~\cite{pareja2020evolvegcn}, 
MTAD-GAT~\cite{zhao2020multivariate}, ROLAND~\cite{you2022roland}. MoE: 
GraphMoE~\cite{huang2025graphmoe}. Baseline scores are averaged 
cross-sectionally and thresholded at validation-set $P_{85}$ under the 
same 7-day detection window.

\noindent\textbf{Implementation.} PyTorch 2.0, NVIDIA A100 (40GB). 
BiLSTM hidden 128, self-attention 4 heads, GAT 8 heads, expert latent 64, 
top-$k=20$, fusion bounds $\alpha_t \in [0.2, 0.8]$, random seed 42. 
Gating: $\tau_{\text{base}}=1.0$, $\beta_{\text{stress}}=0.5$, 
$\mathbf{b}_{\text{div}}=[0.0, 0.5, 0.3, 0.2]$, $\gamma_1=0.01$.

\subsection{Main Results}

Table~\ref{tab:main_results} compares detection performance 
(event detected if MPI $\geq P_{85}$ within a 7-day window; 
lead time = days before event threshold is first crossed). Our 
framework detects all six events with a \textbf{3.7-day mean 
lead time}, outperforming all baselines by \textbf{+33 percentage 
points} over the strongest methods (TranAD, GDN, MTAD-GAT, 
ROLAND, each detecting 4 of 6 events). AUC 0.888 and AP 0.626, 
computed over all 48,300 test-period observations against 
unsupervised proxy labels, substantially exceed all baselines.

The SVB collapse registers 0-day lead time---consistent with its character 
as an abrupt information-driven shock with no pre-event market-wide stress 
buildup, distinguishing it from systemic crises like the Japan unwind 
(4-day lead).

\begin{table}[t]
\centering
\caption{Detection performance on six major financial stress events 
(2023--2024). Lead time averaged over detected events only.}
\label{tab:main_results}
\footnotesize
\setlength{\tabcolsep}{3pt}
\begin{tabular}{@{}lcccc@{}}
\toprule
\textbf{Method} & \textbf{Det. (\%)} & \textbf{Lead (d)} & 
\textbf{AUC} & \textbf{AP} \\
\midrule
\multicolumn{5}{@{}l}{\textit{Temporal Models}} \\
LSTM-AE      & 50.0             & 1.9             & 0.592 & 0.271 \\
TranAD       & \underline{66.7} & 3.0             & 0.669 & 0.373 \\
OmniAnomaly  & 50.0             & 2.4             & 0.625 & 0.304 \\
\multicolumn{5}{@{}l}{\textit{Static Graph}} \\
DOMINANT     & 33.3             & 1.3             & 0.562 & 0.241 \\
GDN          & \underline{66.7} & 2.7             & 0.637 & 0.333 \\
AnomalyDAE   & 50.0             & 2.0             & 0.605 & 0.281 \\
\multicolumn{5}{@{}l}{\textit{Dynamic Graph}} \\
EvolveGCN    & 50.0             & 2.2             & 0.614 & 0.293 \\
MTAD-GAT     & \underline{66.7} & 2.9             & 0.648 & 0.346 \\
ROLAND       & \underline{66.7} & \underline{3.2} & \underline{0.682} 
             & \underline{0.387} \\
\multicolumn{5}{@{}l}{\textit{Mixture-of-Experts}} \\
GraphMoE     & 50.0             & 2.5             & 0.641 & 0.335 \\
\midrule
\textbf{Ours} & \textbf{100.0} & \textbf{3.7} & \textbf{0.888} & 
              \textbf{0.626} \\
\bottomrule
\end{tabular}
\end{table}

\subsection{Ablation Studies}

Table~\ref{tab:ablation} validates each architectural choice. Graph 
fusion requires stress-modulation: both the static prior 
($\mathbf{A}_{\text{prior}}$, AUC 0.712) and purely data-driven graph 
($\mathbf{A}_{\text{learned}}$, AUC 0.651) underperform---adaptive 
interpolation between them is essential. Among the MoE components, 
mechanism-specific feature partitioning matters more than 
stress-aware routing when ablated ($-$0.087 vs.\ $-$0.068 AUC), 
though both contribute independently. Score distributions remain 
stable across all four market regimes (mean $\leq$0.010, P95 drift 
$\leq$3\%), and AUC changes stay $\leq$0.005 across tested 
hyperparameter ranges; inference runs at 50\,ms per timestep on A100, 
comparable to leading baselines.

\begin{table}[t]
\centering
\caption{Ablation results on 2022 validation set.}
\label{tab:ablation}
\footnotesize
\setlength{\tabcolsep}{4pt}
\begin{tabular}{@{}lcc@{}}
\toprule
\textbf{Configuration} & \textbf{AUC} & \textbf{AP} \\
\midrule
\textbf{Full Model} & \textbf{0.871} & \textbf{0.608} \\
\midrule
\multicolumn{3}{@{}l}{\textit{Graph Fusion}} \\
Pure $\mathbf{A}_{\text{prior}}$    & 0.712 & 0.381 \\
Pure $\mathbf{A}_{\text{learned}}$  & 0.651 & 0.334 \\
Fixed $\alpha = 0.5$                & 0.791 & 0.487 \\
\multicolumn{3}{@{}l}{\textit{Expert Specialization}} \\
Single expert    & 0.784 & 0.463 \\
Uniform weights  & 0.803 & 0.491 \\
\bottomrule
\end{tabular}
\end{table}
\subsection{Expert Specialization}

\begin{figure}[t]
\centering
\includegraphics[width=\columnwidth]{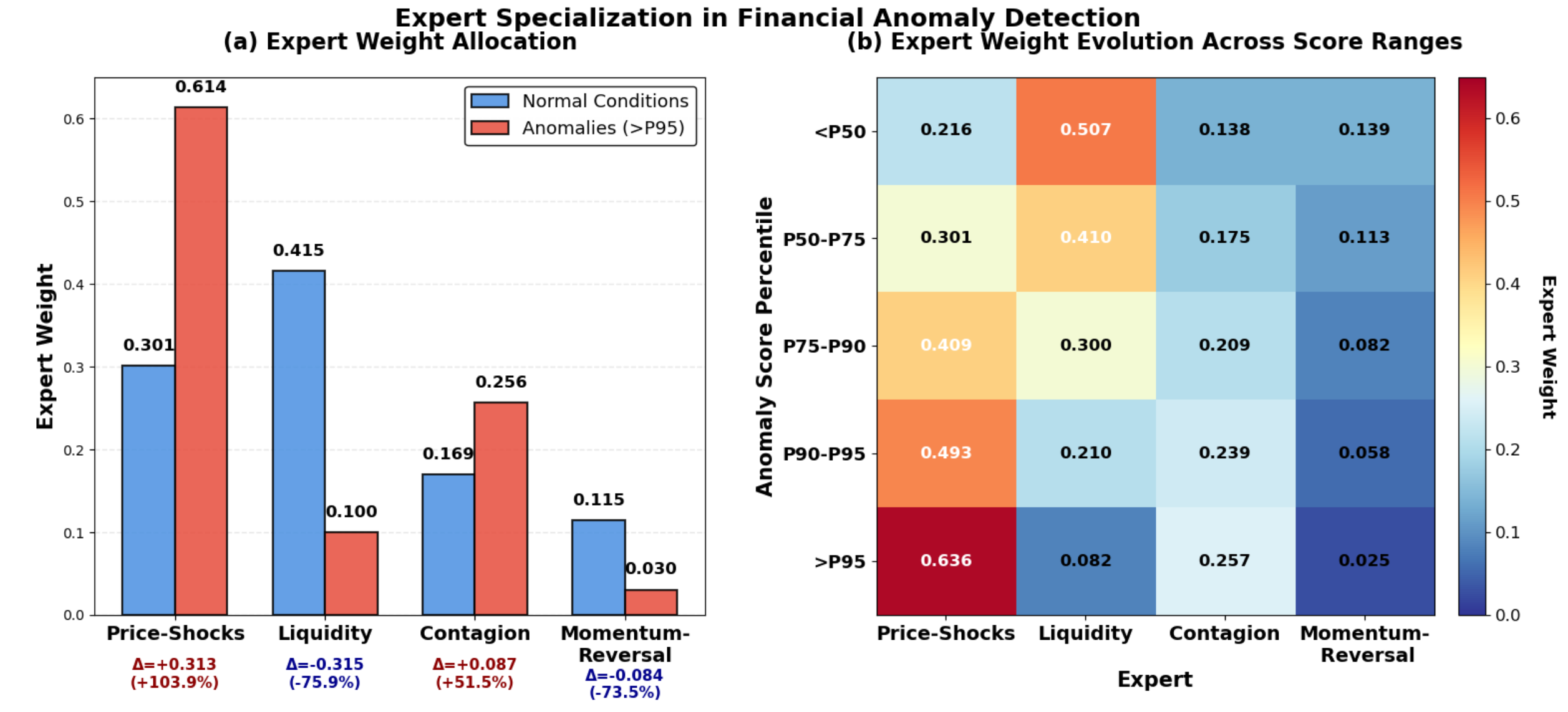}
\caption{\textbf{Learned Routing Weights Shift Toward Financially 
Meaningful Mechanisms as Anomaly Severity Increases.}
\textit{(a)}~Mean routing weights for normal-regime stocks (daily 
bottom-75 by anomaly score) and high-anomaly stocks (daily top-5); 
percentages show shifts relative to normal-regime means.
\textit{(b)}~Mean routing weights across anomaly score percentile bins 
for all test-period stock-day observations: Price-Shock rises 
monotonically and Liquidity falls monotonically with severity. No 
mechanism labels are used during training. Sample: 100 U.S.\ equities, 
2023--2024.}
\label{fig:expert_specialization}
\end{figure}

\noindent\textbf{Expert specialization.}
Financial economics identifies a causal hierarchy among failure 
modes: price shocks trigger, systemic contagion follows as network 
effects take hold, and liquidity deterioration is a downstream 
consequence~\cite{cont2001empirical,elliott2014financial,amihud2002illiquidity}. 
We classify the daily bottom-75 stocks by anomaly score as 
normal-regime and the top-5 as high-anomaly. Figure~\ref{fig:expert_specialization} shows that the model 
rediscovers this ordering from reconstruction error alone---with 
no mechanism labels at any stage: routing shifts strongly toward 
Price-Shock (+104\%) and away from Liquidity ($-$76\%), with 
Systemic-Contagion rising progressively (+52\%), monotonically across 
all severity bins (Price-Shock: 21.6\%$\to$63.6\%; Liquidity: 
50.7\%$\to$8.2\%).

\subsection{Case Studies: Crisis Taxonomy via Routing Dynamics}
\label{sec:experiments-casestudies}

\begin{figure}[t]
\centering
\includegraphics[width=\columnwidth]{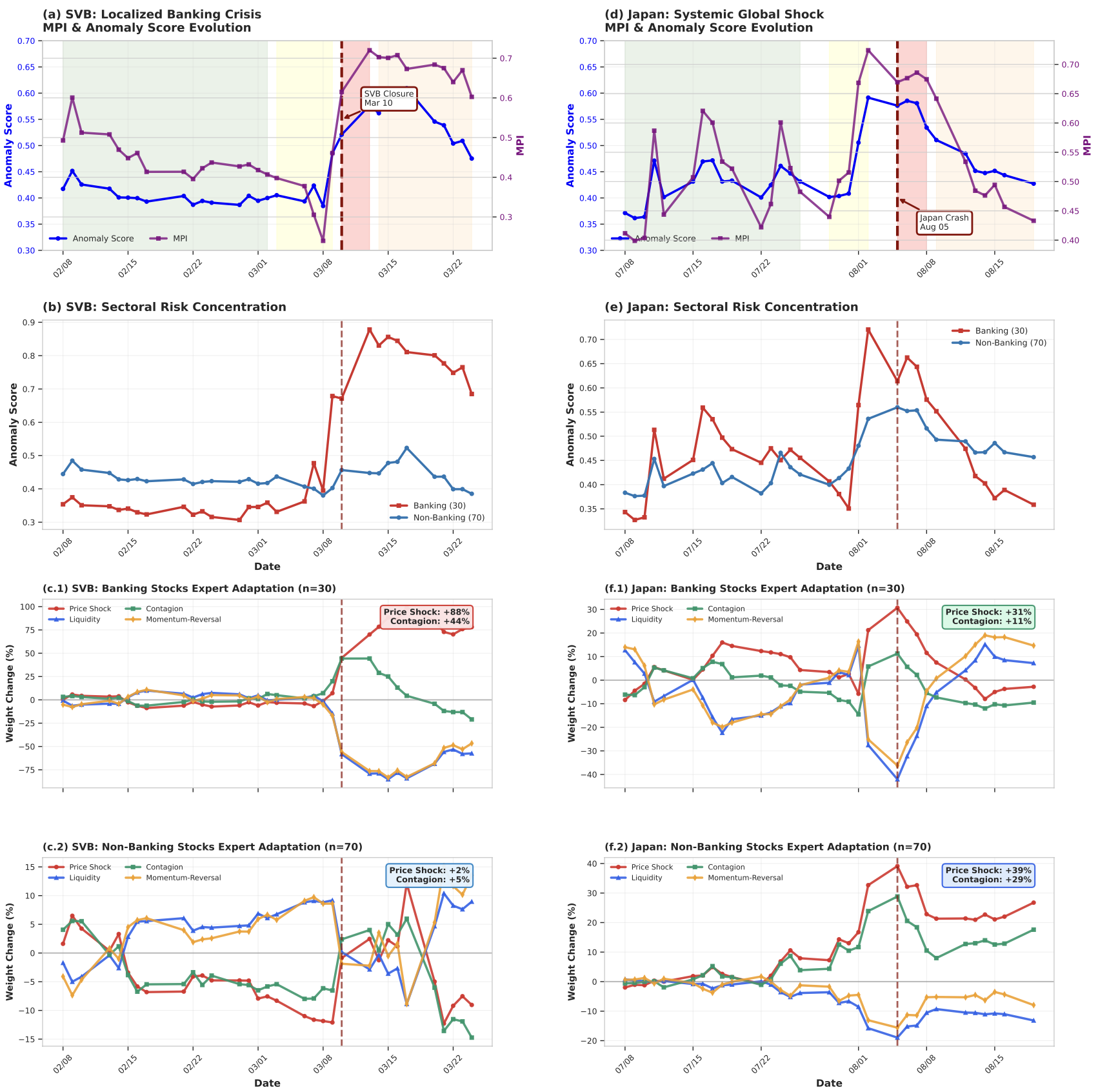}
\caption{\textbf{Routing weight dynamics distinguish crisis type without 
labeled supervision.}
\textit{Left---SVB collapse (March 2023):} MPI and mean anomaly scores 
spike on the event date with no prior elevation (a); daily mean anomaly 
scores for banking ($n=30$) vs.\ non-banking ($n=70$) stocks show 
sharp sectoral divergence at the event date, with banking scores 
surging while non-banking remains near baseline (b); routing weight 
changes are confined to the banking sector, with Price-Shock dominant 
(c.1 vs.\ c.2).
\textit{Right---Japan carry-trade unwind (August 2024):} MPI and mean 
anomaly scores show elevated baselines emerging 1--2 weeks before the 
crash, with MPI crossing the alert threshold 4 days prior (d); daily 
mean anomaly scores rise simultaneously across both banking and 
non-banking stocks, confirming broad market-wide stress (e); routing 
weight changes are symmetric across both sectors (f.1 vs.\ f.2). All 
routing weights are baseline-relative (30--20 days pre-event) and used 
for visualization only.}
\label{fig:case_studies_sector}
\end{figure}

Figure~\ref{fig:case_studies_sector} demonstrates that routing weight 
dynamics yield a natural, supervision-free taxonomy of crisis type. 
We partition the 100 stocks into banking ($n=30$, spanning large-cap, 
regional, and mid-size banks) and non-banking ($n=70$) sectors. Crisis 
scope is quantified via a \textit{sectoral confinement score}---the 
ratio of the dominant mechanism's routing-weight change in the most- 
versus least-affected sector---where scores $>$10 indicate a localized 
crisis and scores $\approx$1 indicate systemic propagation.

\noindent\textbf{SVB collapse (March 10, 2023).} Price-Shock routing 
weights surge in the banking sector (+88\%) while non-banking stocks 
remain near baseline (+2\%), a confinement ratio of 44:1. The 
concurrent Systemic-Contagion activation within banking (+44\%) 
reflects intra-sector propagation to Signature Bank and First 
Republic through shared depositor exposures. MPI first crosses the 
alert threshold on the event date with no prior elevation---an 
abrupt, localized shock rather than gradual systemic accumulation.

\noindent\textbf{Japan carry-trade unwind (August 5, 2024).} The 
Bank of Japan's unexpected rate hike in late July 2024 triggered a 
rapid unwinding of yen carry trades, forcing broad liquidation of 
U.S.\ equity positions across all sectors~\cite{aquilina2024carry}. 
Both banking and non-banking sectors show near-symmetric 
routing-weight activation (confinement ratio $\approx$1:1), a 
signature of systemic global contagion rather than sector-specific 
stress. Non-banking stocks absorb the larger Price-Shock (+39\% vs.\ 
+31\% in banking), reflecting carry-trade investors' concentrated 
leveraged positions in U.S.\ growth and technology equities. The 
model detects the buildup well before the crash: MPI crosses the 
alert threshold 4 days in advance, with elevated anomaly scores 
emerging across both sectors 1--2 weeks prior---actionable warning 
that a localized-crisis detector would miss.

\section{Conclusion}
\label{sec:conclusion}

We presented a mechanism-aware framework for anomaly detection in dynamic 
financial networks that integrates stress-modulated adaptive graph 
fusion, mechanism-aligned mixture-of-experts, and architectural 
interpretability via routing weight dynamics---three capabilities that 
existing methods address only in isolation.

The central finding is that mechanism-aware design delivers 
operational value beyond detection alone: the framework not 
only detects crises earlier than existing methods, but 
characterizes their nature---distinguishing localized sector 
shocks from systemic propagation without labeled supervision. 
This distinction is what makes detection actionable, since identical 
anomaly scores from different mechanisms demand fundamentally 
different interventions. The stability of learned failure modes 
across four heterogeneous market regimes further suggests that 
mechanism-aware architectures generalize in ways that 
regime-specific models cannot, pointing toward a principled 
foundation for transparent, deployable financial risk systems.

\balance
{\footnotesize
\bibliographystyle{IEEEtran}
\bibliography{references}}
\end{document}